\documentclass{IEEEtran}

\usepackage{cite}
\usepackage[export]{adjustbox}

\usepackage{amsmath,amssymb,amsfonts}
\DeclareMathOperator*{\argmax}{argmax} 

\usepackage{algorithm}
\usepackage{algpseudocode}

\usepackage{mathptmx}
\usepackage{graphicx}
\usepackage{textcomp}
\usepackage{xcolor}
\def\BibTeX{{\rm B\kern-.05em{\sc i\kern-.025em b}\kern-.08em
    T\kern-.1667em\lower.7ex\hbox{E}\kern-.125emX}}
\usepackage{makecell}
\usepackage{amsthm}

\usepackage[english]{babel}

\usepackage{caption}
\usepackage{subcaption}

\usepackage{graphicx}
\usepackage{mwe}

\newtheorem{theorem}{Theorem}
\newtheorem{corollary}{Corollary}[theorem]

\usepackage{xcolor}

\usepackage[T1]{fontenc}
\usepackage{etoolbox}

\makeatletter
\patchcmd{\maketitle}{\@fnsymbol}{\@alph}{}{}  
\makeatother

\usepackage[normalem]{ulem}
    
\title{\LARGE \bf
Toward optimal placement of spatial sensors}

\author{
Mingyu Kim$^1$, Harun Yetkin$^{1,2}$, Daniel J. Stilwell$^1$, Jorge Jimenez$^3$, Saurav Shrestha$^4$, and Nina Stark$^4$
\noindent \thanks{$^1$ Mingyu Kim, Harun Yetkin, and Daniel J. Stilwell are with Bradley Department of Electrical and Computer Engineering, Virginia Polytechnic Institute and State University, Blacksburg, VA 24060, USA { \{mkim486, yetkinh, stilwell@vt.edu\}}}
\noindent \thanks{$^2$ Harun Yetkin is also affiliated with the Department of Mechatronics Engineering, Bart{\i}n University, Turkey { \{hyetkin@bartin.edu.tr\} } }%
\noindent \thanks{$^3$ Jorge Jimenez is with Naval Surface Warfare Center, Panama City, FL, USA { \{jorge.g.jimenez2.civ@us.navy.mil\} } } 
\noindent \thanks{$^4$ Saurav Shrestha and Nina Stark are with Charles Edward Via, Jr., Department of Civil and Environmental Engineering, Virginia Polytechnic Institute and State University, Blacksburg, VA 24061, USA { \{sauravs3, ninas@vt.edu\}}}
\noindent \thanks{This work is supported, in part, by the Office of Naval Research via grant N00014-20-1-2845}

}

\begin{document}

\newcommand{\inserttext}[1]{\textcolor{blue}{#1}}
\newcommand{\crosstext}[1]{\textcolor{red}{\sout{#1}}}

\maketitle
\thispagestyle{empty}
\pagestyle{empty}

\begin{abstract}
This paper addresses the challenges of optimally placing a finite number of sensors to detect Poisson-distributed targets in a bounded domain. We seek to rigorously account for uncertainty in the target arrival model throughout the problem. Sensor locations are selected to maximize the probability that no targets are missed. While this objective function is well-suited to applications where failure to detect targets is highly undesirable, it does not lead to a computationally efficient optimization problem. We propose an approximation of the objective function that is non-negative, submodular, and monotone and for which greedy selection of sensor locations works well. We also characterize the gap between the desired objective function and our approximation. For numerical illustrations, we consider the case of the detection of ship traffic using sensors mounted on the seafloor.

\end{abstract}


\section{Introduction}


This paper addresses the challenging task of optimally placing a finite number of sensors to detect Poisson-distributed targets within a bounded domain. The primary objective is to develop an optimal sensor placement algorithm that enables the deployment of sensors based on acquired environmental and target data, possibly allowing for adjustments to sensor locations as new target data becomes available. 


We model target arrivals using a Poisson distribution, and we consider that the target arrival rate, which is represented by the intensity function of the Poisson distribution, is uncertain. To model the uncertainty in the target arrival rate, we employ a log-Gaussian Cox process, which is a Poisson point process where the logarithm of the intensity function is a Gaussian process. We then estimate the underlying intensity function based on prior target arrival data. Based on the estimated intensity function, the selection of sensor locations is determined with the objective of minimizing the probability of failing to detect a target. We show that this objective is equivalent to maximizing the void probability of the Poisson process, which refers to the probability that no targets are undetected. We propose an approximation of the void probability as the objective function for the sensor placement problem. We show that our approximation of the void probability is submodular and monotonic increasing (monotone). Thus, greedy selection of sensor locations works well. For the numerical illustrations, we consider the case of subsea sensors that detect ship traffic. Example ship traffic data is obtained from historical records of the Automated Identification System (AIS) near Hampton Roads Channel, Virginia, USA.

Poisson point processes have been used to model target arrivals in various applications, such as conducting marine mammals surveys~\cite{yuan2017point,jullum2020estimating, makinen2018hierarchical}, disease mapping~\cite{diggle2005point, gatrell1996spatial, cunningham2021quantifying}, crime rate modeling~\cite{shirota2017space}, and border surveillance \cite{szechtman2008models}. The authors in \cite{yuan2017point,jullum2020estimating, makinen2018hierarchical, diggle2005point, gatrell1996spatial, cunningham2021quantifying, shirota2017space} consider a Poisson spatial point process with known intensity function as target arrival model. The authors in \cite{szechtman2008models} assume that target arrivals follow a homogeneous Poisson point process with a known intensity value. In contrast, our approach uses an uncertain intensity function that can be estimated from historical data or in real-time. In~\cite{grant2020adaptive, mutny2021no}, the authors address greedy selection of sensor locations to detect Poisson-distributed target arrivals. However, in these studies, stochasticity in the intensity function is not accounted for. In~\cite{mutny2021sensing}, the authors seek to adaptively identify a stochastic intensity function while choosing a sequence of single observation locations that minimize a reward function related to the number of missed targets. In contrast, we assume that a stochastic intensity function has been identified from historical data, and we seek a set of sensor locations that minimize the probability of that no targets are missed through the entire domain. The existing studies in this field do not analyze the proximity of their solutions to the optimal solution. In our paper, we bridge this gap by conducting an analysis of the deviation between our proposed approximate solution and the optimal solution. 

We model target arrivals as a log-Gaussian Cox process (LGCP) \cite{moller1998log, moller2003statistical, diggle2013spatial}. A Cox process is a Poisson process with a stochastic intensity function. For our applications, we model the intensity function as the log of a Gaussian process. To estimate the intensity function based on prior data, we use Integrated Nested Laplace Approximation (INLA) method, which is a deterministic approximation. INLA approximates the posterior distribution of latent Gaussian models using nested Laplace approximations \cite{rue2009approximate, martino2019integrated, gomez2020bayesian}. We use void probability as our objective function and select sensor locations where the void probability is maximum. We show that in our formulation of the  sensor placement problem, maximizing void probability is the same as minimizing the number of undetected targets.

\subsection*{Contributions}
We address sensor placement using an LGPC target model. Because the optimization problem is numerically challenging, we propose a lower-bound  of the objective function that is submodular and monotone, and for which greedy sensor location selection works well. We further characterize the gap between the desired objective function, which is the probability that no targets are missed, and our lower bound, and we show via numerical examples that the gap appears to be small for representative problems that motivate our analysis.

The organization of the paper is as follows. In Section~\ref{sec:ProblemFormulation}, we present a detection model with multiple sensors and target arrivals modeled as a log-Gaussian Cox process. In Section~\ref{sec:sensorplacement}, we derive a lower bound for the void probability that is submodular and monotone, and facilitates computationally tractable selection of sensors. In Section~\ref{sec:jensengap}, we analyze the gap between void probability and its lower bound from Section~\ref{sec:sensorplacement}. In Section~\ref{sec:numericalresults}, we provide numerical results that show the efficacy of our proposed approach. The appendix shows the proofs of submodularity, monotonicity of the proposed objective function from Section~\ref{sec:sensorplacement}, and of monotonic-decrease of the upper bound of Jensen gap from Section~\ref{sec:jensengap}.

\section{Problem Formulation}
\label{sec:ProblemFormulation}

This paper focuses on the sensor placement problem, specifically addressing scenarios where a set of sensors is used to detect stochastic target arrivals.

\subsection{Sensor model}
We define  $\gamma(s, a_i):S \times S \rightarrow [0,1]$ to be the probability of sensor $i$ detecting a target at location $s$ in a bounded domain $S$ where $a_i$ represents the location of sensor $i$. 

The probability of failing to detect a target at location $s$ with sensor $i$ is expressed $ 1 - \gamma(s, a_i)$. Let $\mathbf{a} = \{a_1, a_2, \ldots, a_M \}$ denote the locations of a set of $M$ sensors. Then, when all $M$ sensors are placed at $\mathbf{a}$, the probability of failing to detect a target at location $s$ is

\begin{equation}
    \pi(s, \mathbf{a}) := \prod_{i=1}^{M}  \big ( 1 - \gamma(s, a_i) \big )
    \label{eq:failingDetectionProb}
\end{equation}

\subsection{Target arrival model: Log-Gaussian Cox Process}
Target arrivals in a bounded region $S$ for a time-interval $T_c$ are modeled by an inhomogeneous Poisson point process with a random intensity $\Lambda(S,T_c)$ where $T_c$ is a time-interval for historical target arrival data collection to compute an estimated target arrival per unit time within the domain. The intensity $\Lambda(S,T_c)$ can be thought of as the expected number of target arrivals in area $S$ over a time interval with length $T_c$ and can be computed

\begin{equation}
\Lambda(S,T_c) = \frac{1}{T_c}\int_{S} \lambda(s) ds \label{eq:intensity}
\end{equation}

\noindent where $\lambda(s):S \rightarrow [0, \infty)$ is the intensity function at location $s \in S$. The intensity function is derived to represent the expected number of targets per unit area in a time-interval $T_c$. We assume that $\lambda(s)$ is stochastic and the logarithm of the spatial variation in the intensity function is a Gaussian process. 

\begin{align}
    \log(\lambda(s)) \sim {GP}(\mu(s), k(s,s')) \label{eq: Gaussian process}
\end{align}

\noindent where $\mu(s)$, $k(s,s')$ are mean and covariance functions respectively and $s', s \in S$. This model is called the log-Gaussian Cox process (LGCP). We refer the reader to \cite{moller1998log, moller2003statistical, diggle2013spatial} for more details on LGCP.  

Given $\Lambda(S,T_c)$, the probability of observing $n$ number targets within $S$ for a time-interval $T$ using Poisson distribution is

\begin{equation*}
    P(N(S,T) = n) = \frac{(\Lambda(S,T_c)T)^{n}}{n!}e^{-\Lambda(S,T_c)T}
\end{equation*}

\noindent where $N(S,T)$ denotes the number of target arrivals.

\section{Suboptimal Sensor Placement}
\label{sec:sensorplacement}
Our goal is to find optimal sensor locations that minimize the number of undetected targets in $S$ and for a time period $T$.

\subsection{Void probability}           

We let $\bar{N}(S,T)$ represent the number of undetected targets in $S$ over time-interval $T$. The probability that $\bar{N}(S,T)$ is zero is computed from the Poisson process

\begin{align}
&P\left(\bar{N}\left(S,T\right) = 0 \mid \lambda(s) \right ) \notag \\
&= \exp\left( - \int_S  \frac{T}{T_c}\lambda(s) \pi(s, \mathbf{a})  ds \right) \label{eq:voidprobabilityPP}       
\end{align}

\noindent where we say that the intensity function $\lambda(s)$ has been \emph{thinned} by the probability of failing to detect a target $\pi(s,\mathbf{a})$. The probability of that $\bar{N}(S,T)$ is zero is known as the \emph{void} probability of the log-Gaussian Cox process. Since we assume the target arrival intensity function $\lambda(s)$ in \eqref{eq:voidprobabilityPP} is stochastic, the void probability is

\begin{align}
      &P\left( \bar{N}\left( S,T \right) = 0 \right) \notag \\
      &=\mathbb{E}_\lambda \biggl[\exp \left(-\int_{S} \frac{T}{T_c}\lambda(s) \pi(s, \mathbf{a}) ds\right)\biggr] \label{eq:thinnedVoidprobability}
\end{align}

\noindent where \eqref{eq:thinnedVoidprobability} represents \eqref{eq:voidprobabilityPP} after marginalizing out $\lambda(s)$.

\subsection{Void probability approximation}

Let $\mathbf{A}$ be the set of all possible sensor locations within $S$ such that the location of a finite number of sensors is $\mathbf{a} \subset \mathbf{A}$. We compute a set of optimal sensor locations such that the void probability of the thinned Cox process is maximized 

\begin{equation}
    \mathbf{a}^\star = \argmax_{\mathbf{a} \subset \mathbf{A}} \mathbb{E}_\lambda  \biggl[\exp \left(-\int_{S} \frac{T}{T_c}\lambda(s) \pi(s, \mathbf{a})  ds\right)\biggr] 
    \label{eq:optimalSensorLocations}
\end{equation}

\noindent The objective function in~\eqref{eq:optimalSensorLocations} is computationally challenging due to a stochastic variable $\lambda(s)$ in the integrand. Therefore, we consider a lower bound for the objective function \eqref{eq:optimalSensorLocations} that can potentially be maximized with less computational effort than directly computing the void probability.

We use Jensen's inequality to obtain a computationally tractable lower bound to \eqref{eq:optimalSensorLocations}. Furthermore, we show that over any discretized set of possible sensor locations, this lower bound is submodular and monotone. Thus, greedy selection of sensor locations is guaranteed to generate sensor locations at which the lower bound is within at least a factor $(1-1/e)$ of the optimal sensor location \cite{krause2008near}. 

Jensen's inequality applied to \eqref{eq:optimalSensorLocations} yields

\begin{align}
    \mathbb{E}_\lambda  \left[ e^{- \tilde{\Lambda}(\mathbf{a})}\right]
    \geq e^{-\mathbb{E}_\lambda [  \tilde{\Lambda}(\mathbf{a})]  }
    \label{eq:jensensinequality}
\end{align}

\noindent where 
\begin{align}
    \Tilde{\Lambda}(\mathbf{a})=\int_{S} \frac{T}{T_c}\lambda(s) \pi(s, \mathbf{a}) ds \label{eq:lambda}
\end{align}
The inequality in \eqref{eq:jensensinequality} provides a lower bound to the void probability. Since the lower bound $e^{-\mathbb{E}_\lambda [  \tilde{\Lambda}(\mathbf{a})]}$ is computationally tractable, we seek a set of sensor location $\Tilde{\mathbf{a}}^\star$ that maximizes the lower bound in \eqref{eq:jensensinequality}. That is    

\begin{align}
    \Tilde{\mathbf{a}}^\star =  \argmax_{\mathbf{a} \subset \mathbf{A}} \ \exp \left( - \int_S \Bar{\lambda}(s)  \pi(s, \mathbf{a})  ds \right) \label{eq:sensorLocations3}
\end{align}

\noindent where we denote $\mathbb{E}_\lambda \bigl[ \frac{T}{T_c}\lambda(s) \bigr]$ by $\Bar{\lambda}(s)$, which is the mean of the intensity function for the Cox process. 

We may apply the logarithm without changing the extremum due to the monotonic nature of the logarithm function. Thus, we may apply the logarithm to the objective function in~\eqref{eq:sensorLocations3}, yielding

\begin{align}
    \Tilde{\mathbf{a}}^\star& =  \argmax_{\mathbf{a} \subset \mathbf{A}} \   - \int_S \Bar{\lambda}(s) \pi(s, \mathbf{a}) ds. 
    \label{eq:meanIntensityObjective}
\end{align}
    
\noindent The objective function in~\eqref{eq:meanIntensityObjective} is submodular and monotonically increasing, but not non-negative. Thus, in order to apply the greedy algorithm to compute a finite number of sensor locations in~\eqref{eq:meanIntensityObjective}, the objective function can be modified by adding a constant term

\begin{align}
    \Tilde{\mathbf{a}}^\star &= \argmax_{\mathbf{a} \subset \mathbf{A}} \  \int_S \Bar{\lambda}(s) ds  - \int_S \Bar{\lambda}(s) \pi(s, \mathbf{a}) ds  
    \label{eq:maxOfObjectiveFunc}
\end{align}

\noindent that yields a non-negative function. 

We compute a set of sensor locations with respect to the objective function in~\eqref{eq:maxOfObjectiveFunc}. Below, we formally address our main result

\begin{theorem}
    The non-negative objective function
    \begin{equation*}
    F(\mathbf{a}) = \int_S \Bar{\lambda}(s) ds - \int_S \Bar{\lambda}(s) \pi(s, \mathbf{a}) ds
    \end{equation*}
    is submodular and monotone. 
    \label{theorem1}
\end{theorem}

\begin{corollary}
    Greedy selection of sensor locations with respect to the objective function in~\eqref{eq:maxOfObjectiveFunc} yields at least $1-1/e$ of the optimal results. 
    \label{corollary1.1}
\end{corollary}

\noindent Proof for Theorem~\ref{theorem1} is in Appendix A. Proof for Corollary~\ref{corollary1.1} follows from Theorem~\ref{theorem1} and is the well-known result in~\cite{nemhauser1978analysis}.

\section{Jensen gap analysis}
\label{sec:jensengap}

Jensen's inequality \eqref{eq:jensensinequality} simplifies the computation of the void probability approximation. However, while this inequality yields a lower bound for the void probability, the lower bound is not necessarily tight. The accuracy of the sensor network we obtain using this approximation depends on the size of this gap, which measures how closely our objective function approximates the void probability. A smaller gap indicates a closer approximation to the void probability. Therefore, the size of the gap is a crucial factor in determining the performance of the sensor network.

In this section, building on the results in~\cite{liao2018sharpening}, we first present an upper bound on the Jensen gap given a sensor placement (Theorem~\ref{theorem2}). Then, by proving that the gap is monotonic decreasing, we show a method to compute the bound (Theorem~\ref{theorem3}). 

\begin{theorem}[Theorem 1 in \cite{liao2018sharpening} ]

    Let $X$ be an one-dimensional random variable with mean $\mu_X$, variance $\sigma_X^2$ and $P(X\in (d_1, d_2))=1$, where $-\infty \leq d_1 \leq d_2 \leq \infty$. Let $\phi(X)$ be a twice-differentiable function on $(d_1,d_2)$. Then, the upper bound of the Jensen gap $J$ is
    \begin{equation*}
    J \leq \sup_{X\in(d_1,d_2)} \left( \frac{\phi(X)-\phi(\mu)}{(X-\mu)^2}-\frac{\phi'(\mu)}{X-\mu} \right)\sigma^2
    \end{equation*}
    \label{theorem2}
\end{theorem}

\noindent In our problem, the random variable $X$ is the expected number of undetected ships $\Tilde{\Lambda}(\mathbf{a}) \in [0,\infty)$ when sensor locations $\mathbf{a}$ are known from \eqref{eq:lambda}. The twice differential function  $\phi(\cdot)$ is $e^{-(\cdot)}$ from the definition of void probability. For simplicity, let the mean and variance of $\Tilde{\Lambda}(\mathbf{a})$ be  $\mu_u$ and $\sigma^2_u$, respectively. Then, Theorem \ref{theorem2} yields
 
\begin{align}
    J\leq J_{up}= \sup_{\Tilde{\Lambda}(\mathbf{a})\in[0,\infty)} \left( \frac{e^{-\Tilde{\Lambda}(\mathbf{a})}-e^{-\mu_u}}{(\Tilde{\Lambda}(\mathbf{a})-\mu_u)^2}+\frac{e^{-\mu_u}}{\Tilde{\Lambda}(\mathbf{a})-\mu_u} \right)\sigma_u^2 \label{eq: upperbound}
\end{align}
\noindent where $J_{up}$ is the upper bound of the Jensen gap.

\begin{theorem}
    When $\mu_u$ and $\sigma_u$ are given from the sensor locations $\mathbf{a}$, the upper bound $J_{up}$ in \eqref{eq: upperbound} is monotonic-decreasing with respect to $\Tilde{\Lambda}(\mathbf{a})$. Therefore, the upper bound of Jensen gap is maximized when $\Tilde{\Lambda}(\mathbf{a})$ is zero, which yields

    \begin{align}
    J_{up}(\sigma_u,\mu_u)=\frac{\sigma_u^2(1-e^{-\mu_u}-\mu_ue^{-\mu_u})}{\mu_u^2} 
    \label{eq: Jensen upper bound when going zero}
    \end{align}
    
    \label{theorem3}
\end{theorem}

\noindent Theorem \ref{theorem3} is based on the fact that expression inside the supremum in \eqref{eq: upperbound} are monotonically decreasing with respect to increasing $\tilde{\Lambda}(\mathbf{a})$, which is proved in Appendix B.

\section{Numerical Results}
\label{sec:numericalresults}

\begin{figure}[t!] 
\centering
\includegraphics[scale=0.41]{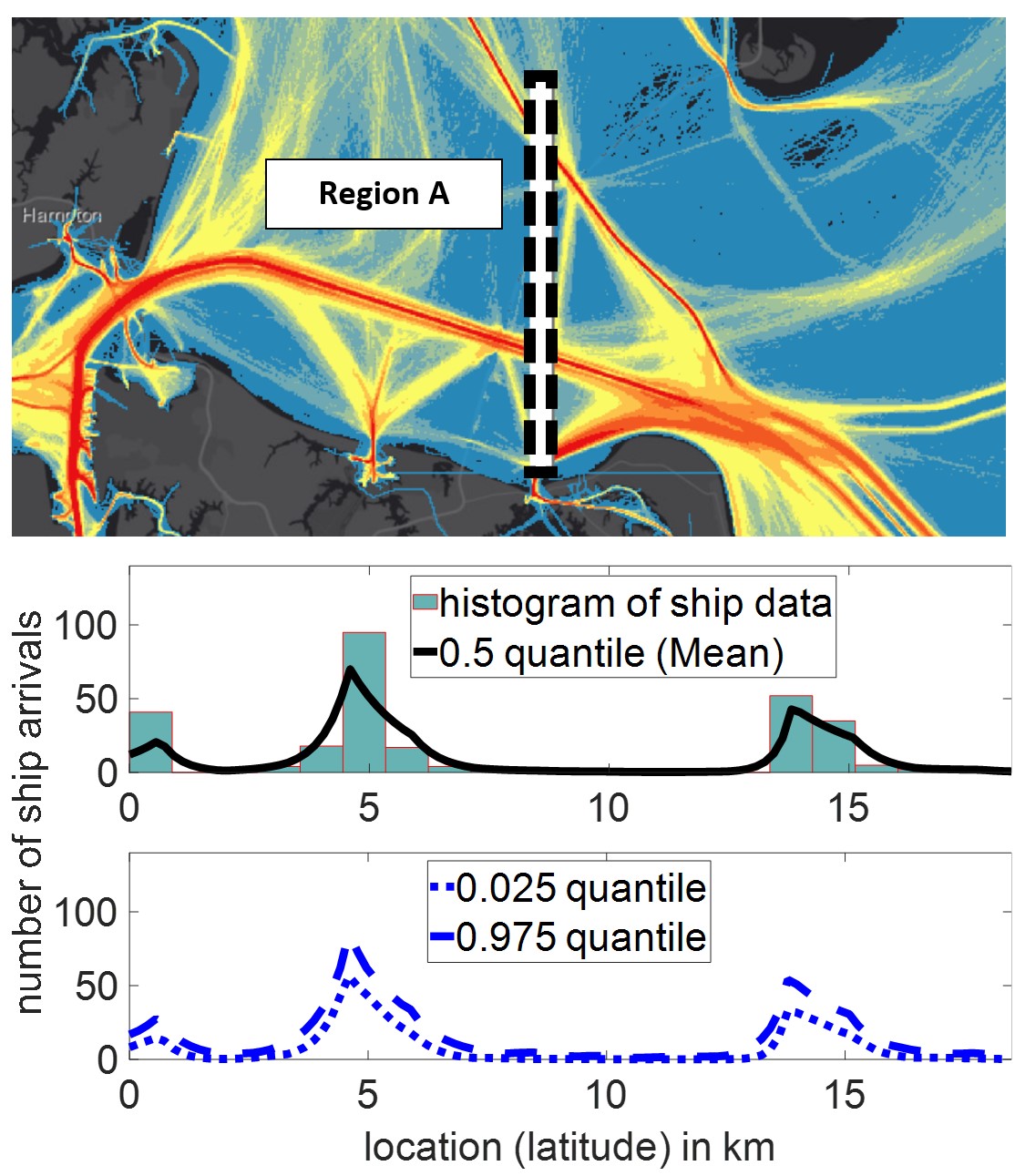}
\caption{(top) Map of Hampton Roads, Virginia, with ship traffic data, water (blue), land (grey). The white bar with the dashed line is the area of interest, Segment A. (middle) The black line is the mean of the estimated intensity function, and the histogram is historical ship traffic data. (bottom) The blue dotted and dashed lines are the lower quantile (2.5\%), upper quantile (97.5\%) of the estimated intensity function.}\label{fig:1}
\end{figure}

In this section, we illustrate our results with numerical examples in which we seek to detect ships using sensors located on the seafloor. We apply INLA to estimate the intensity of ship traffic near Hampton Roads, Virginia. Then we greedily select sensor locations using the objective function in~\eqref{eq:maxOfObjectiveFunc}. Through numerical illustration, we also show that Jensen's gap is small in this example. That is, the difference between the void probability $(\mathbb{E}_{\lambda}[e^{-\Tilde{\Lambda}(\mathbf{a})}]$) and its approximation ($e^{-\mathbb{E}_{\lambda}[\Tilde{\Lambda}(\mathbf{a})]}$) in~\eqref{eq:jensensinequality} is small. We also directly evaluate Jensen's gap for a specific numerical illustration and compare it to the upper bound for Jensen's gap in Section \ref{sec:jensengap}. Our numerical example also shows that the greedy algorithm produces sensor locations that achieve almost the same performance as the optimal sensor locations for the small number of sensors where we can compute optimal locations with respect to void probability via brute force.

We use the ship traffic data near the Hampton Roads Channel, Virginia, USA, provided by the Office for Coastal Management and the Bureau of Ocean Energy Management\cite{marinecadastre.gov}. The data comprises the location (latitude and longitude) of a ship, ship type, and ship detection time. We use the ship traffic data corresponding to the entire month of March 2020 $(=T_c)$, where the domain $S$ in \eqref{eq:intensity} is labeled $A$ in Fig. \ref{fig:1} (top). Region $A$ is treated as one-dimensional line for sensor placement: latitude 36.91676 to 37.08721, longitude -76.08209. Fig.~\ref{fig:1} (top) shows the heat map of ship traffic in the selected area. The red color indicates greater ship traffic has been observed in the area, while the yellow color indicates less ship traffic has been observed, and the blue means no ship has been observed. Within this bounded domain, the possible location where sensors can be placed is discretized with an interval of 50m.

\subsection{Estimation of intensity of ship arrival model}

In order to estimate the intensity function, we use the inlabru package in R~\cite{bachl2019inlabru}, which builds on the R-INLA package~\cite{lindgren2015bayesian}. We consider a zero mean Gaussian process with a Matern covariance function 
\begin{equation*}
    k(s,s^\prime)= \sigma_u^2 \ \frac{2^{1-\zeta}}{\Gamma(\zeta)} \ (\kappa ||s - s^\prime || )^{\zeta} \ K_{\zeta} \ (\kappa ||s - s^\prime ||) 
\end{equation*}
\noindent where $s$ and $s^\prime$ are two locations within the domain, $\sigma_u$ is the variance, $\zeta > 0$ is the smoothness parameter, $\kappa=\sqrt(8\zeta)/\beta >0$ is the scale parameter, $|| \cdot ||$ denotes the Euclidean distance, $K_\zeta$ is the modified Bessel function of second kind, and $\beta$ is a spatial range parameter (see~\cite{rpackagedocumentation_2019} for more details). 
We use the following parameter values for the numerical illustrations: $\zeta = 1.5$, $\beta, \sigma_u$ from $P(\beta<\beta_0 = 150)= 0.75$, $P(\sigma_u > \sigma_{u0} = 0.1) = 0.75$, respectively. As shown in Fig. \ref{fig:1} (middle, bottom), with the parameters, covariance function above, and historical ship traffic data in March 2020 (histogram), we estimate the mean (black line) and the 95$\%$ of the confidence interval (blue lines) using INLA.

\subsection{Sensor model}
For the probability of sensor $i$ detecting a ship, we use the sensor model

\begin{equation}
    \gamma(s, a_i) = \rho e^{-(a_i-s)^2/\sigma_l}
    \label{eq:probabilityOfDetection}
\end{equation}

\noindent where $0 \leq \rho \leq 1$ is the maximum probability of detection and $\sigma_l$ is the length scale parameter. For the numerical illustrations, we consider that $\rho = 0.95 $ and $\sigma_l = 0.9 $.

\subsection{Sensor placement for maximizing void probability approximation}

We seek to maximize void probability directly, 
 (probability that number of undetected ships is zero), but instead, we select sensor locations using the lower bound for void probability in Section \ref{sec:sensorplacement}. Furthermore, we evaluate the difference between void probability and its lower bound. We do not directly consider optimal sensor location selection. Rather, we evaluate the utility in this numerical illustration of greedily selecting sensor locations to maximize the lower bound. Furthermore, we evaluate the difference between greedy and optimal selection of sensor locations when maximizing the lower bound for small numbers of sensors for which we can compute optimal sensor locations using brute force.

\begin{figure}[t!] 
\centering
\includegraphics[scale=0.35]{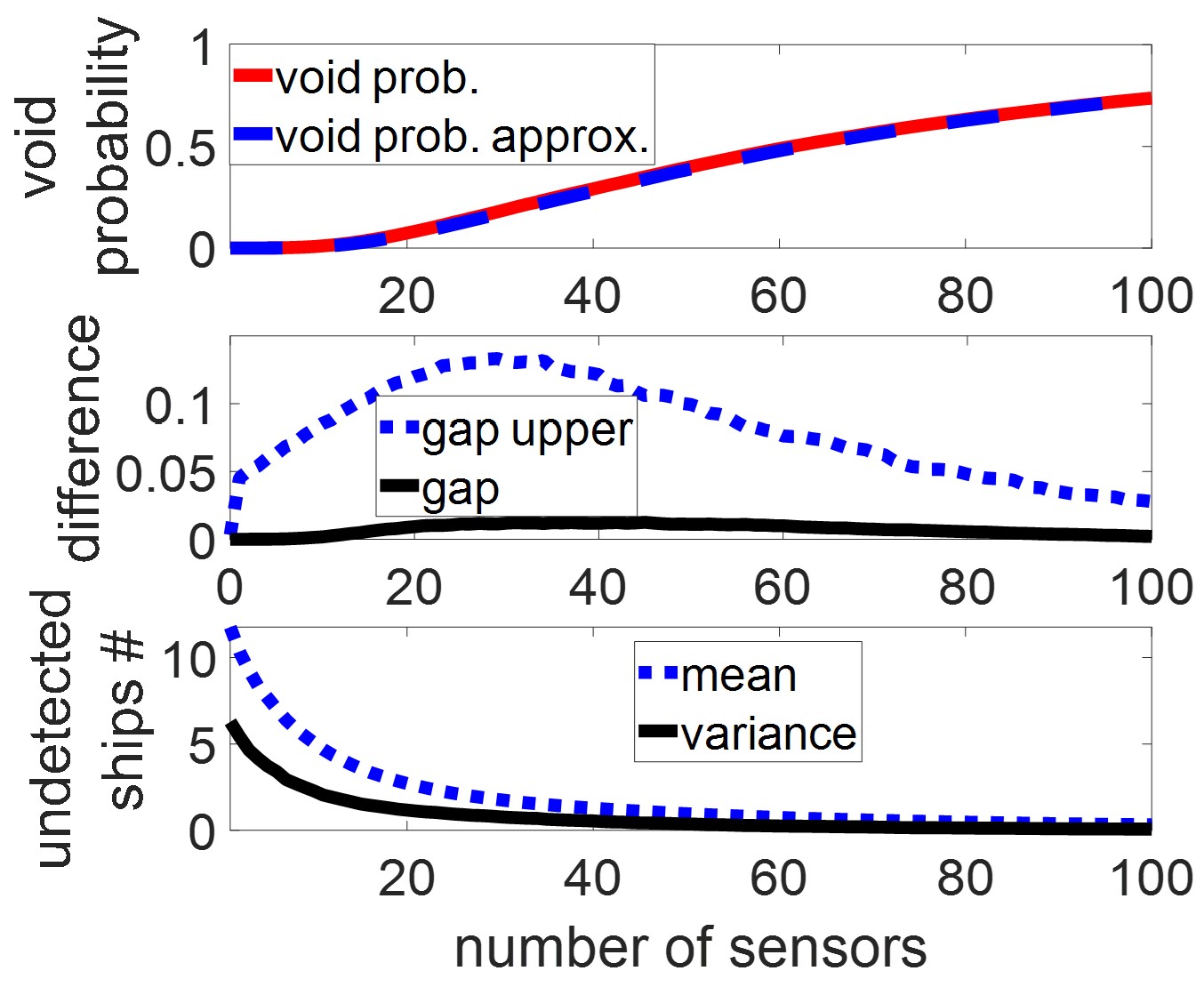}
\caption{(top) With March 2020 data of Segment A, showing void probability (red line) and void probability approximation (blue dashed line) as greedily placing 100 sensors. (middle) Actual difference (black line = Jensen gap) between void probability and void probability approximation, and Jensen gap upper bound (blue dotted line) using \eqref{eq: Jensen upper bound when going zero}. (bottom) Mean (blue dotted line) and variance (black line) of the number of undetected ships} \label{fig:2}
\end{figure}

Given the estimated intensity function and the probability of detection from~\eqref{eq:probabilityOfDetection}, we compute the suboptimal sensor locations. Fig.~\ref{fig:2} (middle) shows Jensen's gap, which is the difference between void probability ($\mathbb{E}_{\lambda}[e^{-\Tilde{\Lambda}(\mathbf{a})}]$) and void probability approximation ($e^{-\mathbb{E}_{\lambda}[\Tilde{\Lambda}(\mathbf{a})]}$). For the results in Fig.~\ref{fig:2}, using the objective function in~\eqref{eq:maxOfObjectiveFunc}, we first greedily compute the suboptimal sensor locations that maximize the void probability approximation. We evaluate the void probability for the suboptimal sensor locations using Monte Carlo method. We sample a large number ( $\geq 10,000$) of the ship arrival intensity functions $\hat{ \lambda}_j$ from \eqref{eq: Gaussian process}, which has been estimated by using INLA. The average void probability for the greedily selected sensor locations is 
\begin{align}
    \frac{\sum_{j}^{W_{\lambda}}\exp \left(-\int_{S} \hat{\lambda}_{j}(s) \prod_{i=1}^{M} \big ( 1 - \gamma(s, \Tilde{a}_{i}^{\star})  \big ) ds\right)}{W_{\lambda}} \label{eq: averagevoid}
\end{align}
where $W_{\lambda}$ is the number of Monte Carlo sampled functions of the stochastic estimated intensity function of ship arrival $\lambda(s)$. Correspondingly, the $j^{th}$ sampled function is $\hat{\lambda}_{j}(s)$, and $\Tilde{a}_{i}^{\star}\in \Tilde{\mathbf{a}}^{\star}(=\{\Tilde{a}_1^{\star},...,\Tilde{a}_M^{\star} \})$ is $i^{th}$ greedily selected sensor location. For simplicity, we denote $\hat{\lambda}_{j}(s)=\frac{T}{T_c}\lambda_{j}(s)$. Fig. \ref{fig:2} (top) shows the void probability approximation (dashed blue line) with the void probability (red line) for the same set of sensor locations. This process is repeated for the number of sensors $M$ varying from 0 to 100.

\begin{table}[t!]\begin{center}
\caption{Comparison of Fig.\ref{fig:2} results for the void probability and void probability approximation from  \eqref{eq:jensensinequality} after placing 100 sensors.\label{tab:tab1}}
\begin{tabular}{c|c|c}
\hline
  \thead{} & 
  \thead{Computation time\\(sec)}  &  
  \thead{Worst percentage \\ difference (\%) }\\
\hline
\thead{VP (void prob.) }    & 150715.67     &   0     \\ 
\thead{VP approx. }        & 0.068        &   1.77  \\
\hline
\end{tabular}

\end{center}
\end{table}

\begin{table}[t!]\begin{center}
\caption{Comparison of Fig.\ref{fig:3} results for greedy selection and optimal search.\label{tab:tab2}}
\begin{tabular}{c|c|c|c}
\hline
 \thead{Number of \\ sensors} &
 \thead{\% of greedy VP \\ of optimal VP} &  
 \thead{Computation time \\ for greedy (sec)}& 
 \thead{Computation time \\ for optimal (sec)}\\
\hline
$2$ & 100       &   0.013  &  2.45    \\ 
$3$ & 100       &   0.015  &  10.46   \\
$4$ & 100       &   0.016  &  1516.61    \\
$5$ & 98.29  &   0.017  &  245257.43    \\ 
\hline
\end{tabular}
\end{center}
\end{table}


\begin{figure}[t!]
\centering 
\includegraphics[scale=0.37]{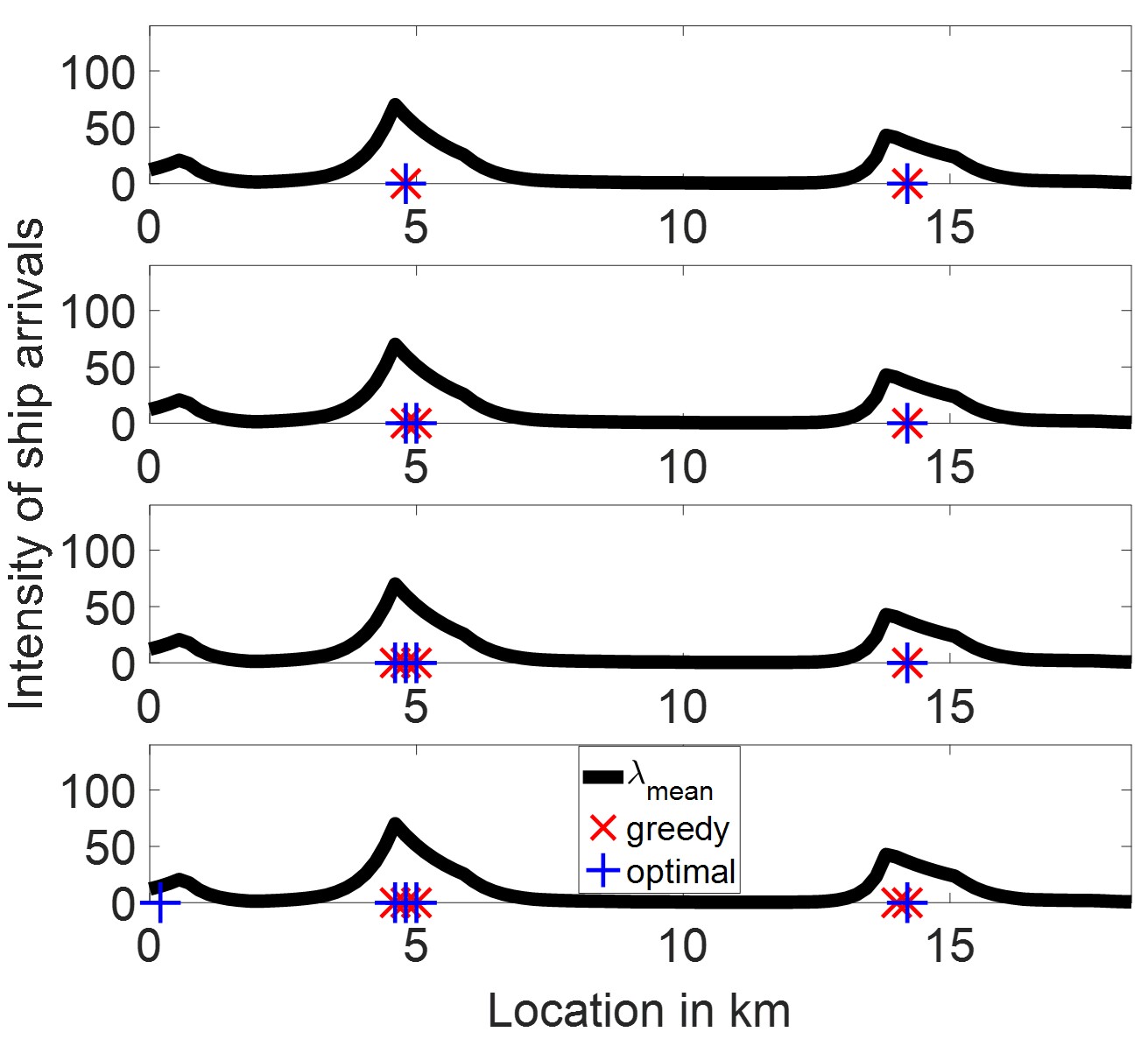}
\caption{Comparison of sensor placement for greedy selection and optimal case using the objective function sensor~\eqref{eq:maxOfObjectiveFunc}. Red cross signs are greedy sensor placement. Blue plus signs are optimal sensor placement. Placing 2, 3, 4, and 5 sensors, respectively, from top to bottom.}
\label{fig:3}
\end{figure}

As shown in Fig.~\ref{fig:2} (middle), the maximum percent difference between the void probability and its approximation is less than 0.0125, and as we place more sensors, the gap tends to be smaller. As discussed in Sec. \ref{sec:jensengap}, using \eqref{eq: Jensen upper bound when going zero}, the upper bound of Jensen gap is computed with the expected value of undetected number of ships $\mu_u$ and its variance $\sigma_u^2$ shown in Fig.~\ref{fig:2} (bottom). As shown as a blue dotted line in Fig.~\ref{fig:2} (middle), the maximum upper bound of Jensen gap is approximately 0.15 and the Jensen gap (black line) is less than or equal to the upper bound. Table \ref{tab:tab1} shows that while the computation time for greedily placing 100 sensors is less than 0.1 seconds, evaluating the void probability at the same locations takes 150715.67 seconds. 


\subsection{Small number of sensor placement for void probability}
Fig.~\ref{fig:3} shows sensor location for both greedy and optimal sensor placement for the number of sensors varying from 2 to 5. Correspondingly, Table~\ref{tab:tab2} shows a comparison of the performance of the greedy selection and the optimal sensor placement. It demonstrates that the greedy selection performs well compared to the optimal. In our numerical experiment, the algorithms are implemented in MATLAB on a Windows computer that has a processor of Core i7 CPU with 1.3 GHz and a RAM of 16.0 GB.

\section{Conclusion}
\label{sec:conclusion}

We propose a computationally tractable suboptimal sensor placement method using void probability approximation as an objective function. This proposed objective function takes into account a stochastic target arrival intensity function. We show that the modified void probability approximation is non-negative, submodular, and monotone, which allows us to use the greedy selection method. Furthermore, we analyze Jensen gap and provide an upper bound for Jensen gap. In numerical illustrations, with historical ship traffic data, we demonstrate that a greedy algorithm for choosing sensor locations yields suboptimal results. 

\section*{Appendix A: proof of submodularity and monotonicity}

$\textit{Proof}: $
Let $F(\mathbf{a})$ be defined as
\begin{align}
    F(\mathbf{a}) = \int_S \Bar{\lambda}(s) ds  -\int_S \Bar{\lambda}(s)\pi(s, \mathbf{a}) ds \notag 
\end{align}

\noindent where $\Bar{\lambda}(s)$ is the non-negative expectation of intensity function and $\pi(s, \mathbf{a})$ is defined in~\eqref{eq:sensorLocations3}. For the location of the set of sensors $A, B, C$ such that $A \subseteq B \subset C$ and for a new common sensor location (of $A,B$) $\hat{a}\in C \backslash B$, $F(\mathbf{a})$ is submodular if the following inequality holds
\begin{align}
    F(A\cup \{\hat{a}\}) - F(A) \geq F(B\cup \{\hat{a}\}) - F(B)
    \label{eq:submodularityCondition}
\end{align}

For $\pi(s,A)$ and $\pi(s,B)$, the sensor network $A$ and $B$ are composed of location of $M_1$ and $M_2$ sensors ($M_1 \leq M_2$) respectively. Then, $\pi(s,A)$ is

\begin{align*}
    \pi(s,A) &=  \prod_{i=1}^{M_1}(1-\gamma(s,a_i))
\end{align*}

\noindent Then, similarly with the set of $B$, for  $\pi(s,B)$
\begin{align*}
    \pi(s,B) &=  \prod_{i=1}^{M_2}(1-\gamma(s,a_i))
\end{align*}

\noindent With the common sensor location $\hat{a}$,
\begin{align*}
    \pi(s,A\cup \{\hat{a}\}) &= \pi(s,A)(1-\gamma(s,\hat{a})) \\
    \pi(s,B\cup \{\hat{a}\}) &= \pi(s,B)(1-\gamma(s,\hat{a}))
\end{align*}
With the modified objective function in~\eqref{eq:maxOfObjectiveFunc} 
\begin{align*}
    F(\mathbf{a}) &=\int_S \Bar{\lambda}(s) ds-\int_S \Bar{\lambda}(s) \pi(s,\mathbf{a}) ds
\end{align*}
such that 
\begin{align*}
    F(A) &=\int_S \Bar{\lambda}(s) ds-\int_S \Bar{\lambda}(s) \pi(s,A) ds\\
    F(B) &=\int_S \Bar{\lambda}(s) ds-\int_S \Bar{\lambda}(s) \pi(s,B) ds\\
    F(A\cup \{\hat{a}\}) &=\int_S \Bar{\lambda}(s) ds-\int_S \Bar{\lambda}(s) \pi(s,A\cup \{\hat{a}\})ds\\
    F(B\cup \{\hat{a}\}) &=\int_S \Bar{\lambda}(s) ds-\int_S \Bar{\lambda}(s) \pi(s,B\cup \{\hat{a}\}) ds\\
\end{align*}
Then, as long as \eqref{eq:submodularityCondition} holds, $F(\mathbf{a})$ is submodular. The left (LHS) and right-hand side (RHS) of the inequality \eqref{eq:submodularityCondition} are
\begin{align}
    F(A\cup\{\hat{a}\})-F(A) &= \int_S \Bar{\lambda}(s) (\pi(s,A)-\pi(s,A\cup \{\hat{a}\})) ds \notag\\
    F(B\cup\{\hat{a}\})-F(B) &= \int_S \Bar{\lambda}(s) (\pi(s,B)-\pi(s,B\cup \{\hat{a}\})) ds \notag
\end{align}

\noindent By subtracting RHS from LHS 
\begin{align}
    (F(A\cup\{\hat{a}\})&-F(A))-(F(B\cup\{\hat{a}\})-F(B)) \notag\\ 
    =\int_S& \Bar{\lambda}(s) [(\pi(s,A)-\pi(s,A\cup \{\hat{a}\})) \notag\\
    & -(\pi(s,B)-\pi(s,B\cup \{\hat{a}\}))] ds\notag\\
    =\int_S&\Bar{\lambda}(s) \pi(s,A)\pi(s,\hat{a}) \times
     \notag \\ &\left(1-\prod_{j=M_1+1}^{M_2}\left(1-\gamma(s,a_j)\right)\right)  ds \label{submodular}
\end{align}
where $\pi(s,\hat{a}) = 1-\gamma(s,\hat{a})$.
In \eqref{submodular}, the result consists of non-negative four components: $\Bar{\lambda}(s)$ is non-negative and $\hat{\pi}(M_1,t)$, $\hat{\pi}(\hat{a},t)$, and the rest of the term are between zero and one. Therefore, $(\pi(s,A)-\pi(s,A\cup \{\hat{a}\}))-(\pi(s,B)-\pi(s,B\cup \{\hat{a}\})) \geq 0$. 
That is, $F(A\cup\{\hat{a}\})-F(A) \geq F(B\cup\{\hat{a}\})-F(B)$. Therefore, it proves that $F(\mathbf{a})$ where $\mathbf{a}=\{ a_1,...,a_M\}, a_i \in S $ is non-negative submodular. 

To prove that $F(\mathbf{a})$ is monotonic increasing, we show that the $F(A) \leq F(B)$ holds. By subtracting $F(A)$ from $F(B)$


\begin{align}
          F(B) - F(A) = \int_S \Bar{\lambda}(s) (\pi(s,A)-\pi(s,B)) ds \nonumber 
\end{align}

\noindent Due to the fact that $\pi(s,A)$ is greater than or equal to $\pi(s,B)$ and $\Bar{\lambda}(s)$ is non-negative, $0 \leq F(B)-F(A) $. That is equivalent to $F(A) \leq F(B)$. Thus, $F(\mathbf{a})$ is monotonic increasing
\qedsymbol.


\section*{Appendix B: proof of monotonic-decrease of $J_{up}$}
$\textit{Proof}: $ We can rewrite \eqref{eq: upperbound} as
\begin{align*}
        J_{up}&= \sup_{\Tilde{\Lambda}(\mathbf{a})\in[0,\infty)}  \frac{\sigma_u^2\left(e^{-\Tilde{\Lambda}(\mathbf{a})}-e^{-\mu_u}+\Tilde{\Lambda}(\mathbf{a})e^{-\mu_u}-\mu_ue^{-\mu_u}\right)}{(\Tilde{\Lambda}(\mathbf{a})-\mu_u)^2} \\
        &=\sup_{\Tilde{\Lambda}(\mathbf{a})\in[0,\infty)}  \frac{\sigma_u^2 e^{-\mu_u}\left(e^{-(\Tilde{\Lambda}(\mathbf{a})-\mu_u)}-1+\Tilde{\Lambda}(\mathbf{a})-\mu_u\right)}{(\Tilde{\Lambda}(\mathbf{a})-\mu_u)^2}
\end{align*}
Let $y=\Tilde{\Lambda}(\mathbf{a})-\mu_u \in [-\mu_u,\infty)$. Then, the upper bound is
\begin{align*}
        &=\sup_{y\in[-\mu_y,\infty)}  \frac{\sigma_u^2 e^{-\mu_u}\left(e^{-y}-1+y\right)}{y^2}
\end{align*}

\noindent Given $\mu_u$ and $\sigma_u^2$, if $h(y)=\frac{e^{-y}-1+y}{y^2}$ is monotonic-decreasing, $J_{up}$ is monotonic-decreasing. The function $h(y)$ is monotonic-decreasing if 
\begin{align}
    \frac{\partial h(y)}{\partial y}=\frac{(2-y)-e^{-y}(y+2)}{y^3} \leq 0  \label{eq: monotonicdecreasing}
\end{align}

\noindent There are a number of ways to show that \eqref{eq: monotonicdecreasing} is satisfied. One approach is to evaluate the roots of the numerator as a polynomial in $y$, and show that the roots are not real, and thus the numerator does not change sign. The sign of \eqref{eq: monotonicdecreasing} is then evaluated separately for the case that $y>0$ and $y<0$ due to the fact that the denominator changes the sign. For the case that $y = 0$, application of L'Hopital's rule twice shows that the ratio remains well defined at $y=0$ \qedsymbol.

\IEEEtriggeratref{11}
\bibliographystyle{IEEEtran}
\bibliography{main.bib}

\end{document}